\documentclass[runningheads]{llncs}
\usepackage{graphicx}
\usepackage{amsfonts,amssymb} 
\usepackage{bm}
\usepackage{makecell}

\begin{document}
\title{Multi-Granularity Framework for Unsupervised Representation Learning of Time Series}
%
%
\author{Chengyang Ye \inst{1} \orcidID{0000-0002-3706-8805} \and
Qiang Ma \inst{2} \orcidID{0000-0003-3430-9244}}
%
%
\institute{$^1$ Graduate School of Informatics, Kyoto University, Kyoto, Japan\\
$^2$ Graduate School of Science and Technology, Kyoto Institute of Technology, Kyoto, Japan\\
\email{ye.chengyang.67x@st.kyoto-u.ac.jp}\\
\email{qiang@i.kyoto-u.ac.jp}}
\maketitle              
\begin{abstract}
Representation learning plays a critical role in the analysis of time series data and has high practical value across a wide range of applications. including trend analysis, time series data retrieval and forecasting. In practice, data confusion is a significant issue as it can considerably impact the effectiveness and accuracy of data analysis, machine learning models and decision-making processes. In general, previous studies did not consider the variability at various levels of granularity, thus resulting in inadequate information utilization, which further exacerbated the issue of data confusion. This paper proposes an unsupervised framework to realize multi-granularity representation learning for time series. Specifically, we employed a cross-granularity transformer to develop an association between fine- and coarse-grained representations. In addition, we introduced a retrieval task as an unsupervised training task to learn the multi-granularity representation of time series. Moreover, a novel loss function was designed to obtain the comprehensive multi-granularity representation of the time series via unsupervised learning. The experimental results revealed that the proposed framework demonstrates significant advantages over alternative representation learning models.

\end{abstract}
\section{Introduction}
Time series is a traditional and important type of data that is ubiquitous in numerous fields. Significant progress in the widespread use of sensors and social production activities has further promoted the development of time series data such as electrocardiograms (ECG) \cite{tseng2020healthcare} and daily stock prices \cite{lawi2022implementation}. With the development of machine learning and data mining, representation learning, which can reveal hidden information in time series by establishing high-dimensional representations, has been increasingly applied to the field of time series. 

However, despite the recent challenges and advancements made by deep learning models in tasks such as prediction and classification, the dominant position of representation learning methods in time series has yet to be established, in contrast to fields such as computer vision (CV) \cite{parvaiz2023vision} and natural language processing (NLP) \cite{zhou2023comprehensive}. In particular, non-deep learning methods, such as HIVE-COTE \cite{lines2018time} and TS-CHIEF \cite{shifaz2020ts}, provide unique advantages. 

\begin{figure}[!tb]
\includegraphics[width=\textwidth]{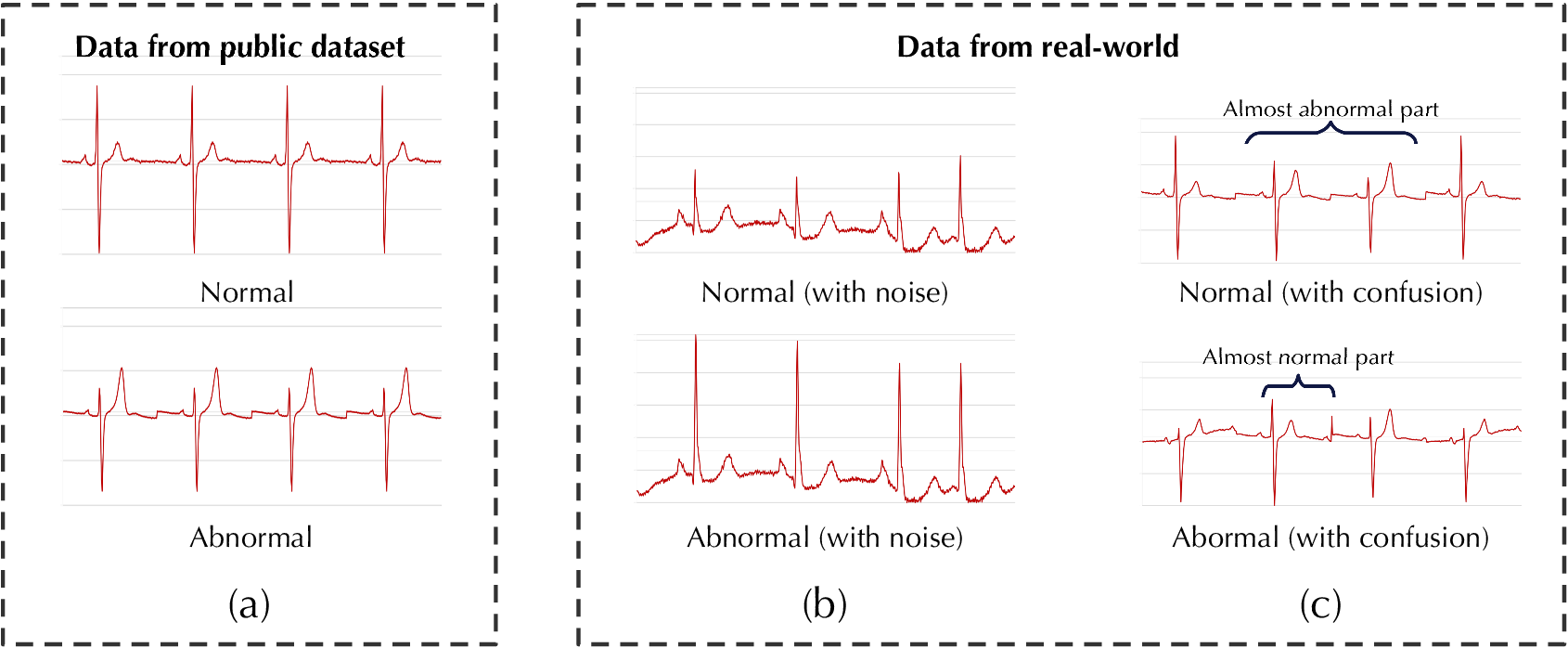}
\caption{Example of ECG data from public dataset and real-world. These figures present several issues of data quality in real-world ECG data.} \label{fig1}
\end{figure}

Although multiple time series representation methods achieve adequate results on public datasets, in real-world application scenarios, time series data are generally subject to missing data, noise data, and data confusion, among other adverse conditions. Fig. 1 presents a typical example of this issue in ECG data. In public datasets designed for model training, the ECG data contain more typical class features (normal and abnormal), without noise or confusion (as shown in Fig. 1(a)). However, in practice, the time series data quality is different. Noise is a common problem in real-world applications, as illustrated in Fig. 1(b), which can have several negative effects on data analysis, including reduced accuracy and misleading conclusions. In addition, data confusion is a more significant issue, as it can considerably impact the effectiveness and accuracy of data analysis. Data confusion refers to cases wherein data from different categories, sources, or contexts is mixed or entangled, thus making it difficult to discern clear patterns, relationships, or structures within the data. Considering the ECG data as an example, overlapping or ambiguous morphologies frequently appear in real-world data (as shown in Fig. 1(c)). Electrocardiograms data from different cardiac conditions may exhibit similar or overlapping morphologies, which makes it challenging to distinguish between them. For example, certain types of arrhythmias may appear similar to a normal sinus rhythm, thus leading to data confusion. 

To address this issue, multiple studies comprehensively considered representations of time series at different granularity, i.e., multi-granularity methods \cite{reis2019multiscale}. An simple example of a multi-granularity method is sales reports that includes data at both the individual transaction and aggregate levels such as monthly or yearly totals. By capturing information from multiple scales or levels of detail, these approaches improve the robustness and accuracy of the analysis and interpretation of time series data. Although multi-granularity representations provide more information, information redundancy is generally observed between different granularities. This redundancy can potentially lead to increased computational complexity, and render the analysis and interpretation of time series data more challenging. Moreover, numerous existing multi-granularity methods are focused primarily on the simple fusion of decision results, and generally require the re-design of representation models. Consequently, they cannot utilize existing, well-performing representation methods and lack the flexibility to adapt to different scenarios.

This paper proposes a novel unsupervised learning framework named MUG (for \textbf{MU}lti-\textbf{G}ranularity), which combines the multi-granularity features of time series based on existing representation learning research. The proposed general framework integrates two different granularities of time series representation methods: a fine-grained representation method, which represents timestamp-level time series data, and a coarse-grained representation method, which represents segment-level time series data. Specifically, for the multiple fine-grained time series representation results, we employed a vector fusion method based on attention mechanism to obtain a comprehensive representation. In addition, based on multi-modal fusion techniques, we employed a cross-granularity attention mechanism to map of coarse-grained representations onto fine-grained representations. This allowed for the fusion of the overall features in the coarse-grained representations with the detailed information in the fine-grained representations. Finally, based on the retrieval task, we designed a more suitable training method for the multi-granularity time series representation learning. 

The main contributions of this study are as follows:

\begin{itemize}
\item[-]This paper presents a focused study on the transformer-based fusion model of multi-granularity representation for time series data. In particular, this paper proposes a novel unsupervised learning framework (Section 3.1) to build association between timestamp-level and segment-level features.   
\item[-]We developed an unsupervised training method (Section3.3). In particular, a retrieval task for the time series data with a unique loss function was designed to obtain the comprehensive multi-granularity representation of time series via unsupervised training. 
\item[-]We conducted extensive experiments on several public datasets from different fields and real-world datasets (Section 4). In comparison with other baseline algorithms, the proposed MUG model achieved an improved performance. 
\end{itemize}

The remainder of this paper is organized as follows: Section 2 outlines previous studies on representation learning for time series, in addition to multi-granularity representation methods for time series from the existing literature. Section 3 presents the architecture of the proposed framework in detail. Thereafter, Section 4 presents the experimental results, followed by a summary of conclusions in Section 5.

\section{Related Work}

\subsection{Representation Learning of Time Series}
The representation learning of time series data has attracted considerable research attention in recent years. The primary objective of these models is to identify spatio-temporal dependencies in the data, which can help uncover the underlying patterns, trends, and relationships that can be used for various tasks, such as forecasting, classification, and anomaly detection. 

According to representation granularity, the existing representation learning models of time series can be broadly classified into two categories: coarse- and fine-grained representation methods. The differences between the two types are shown in Fig. 2. 

\begin{figure}[!tb]
\includegraphics[width=\textwidth]{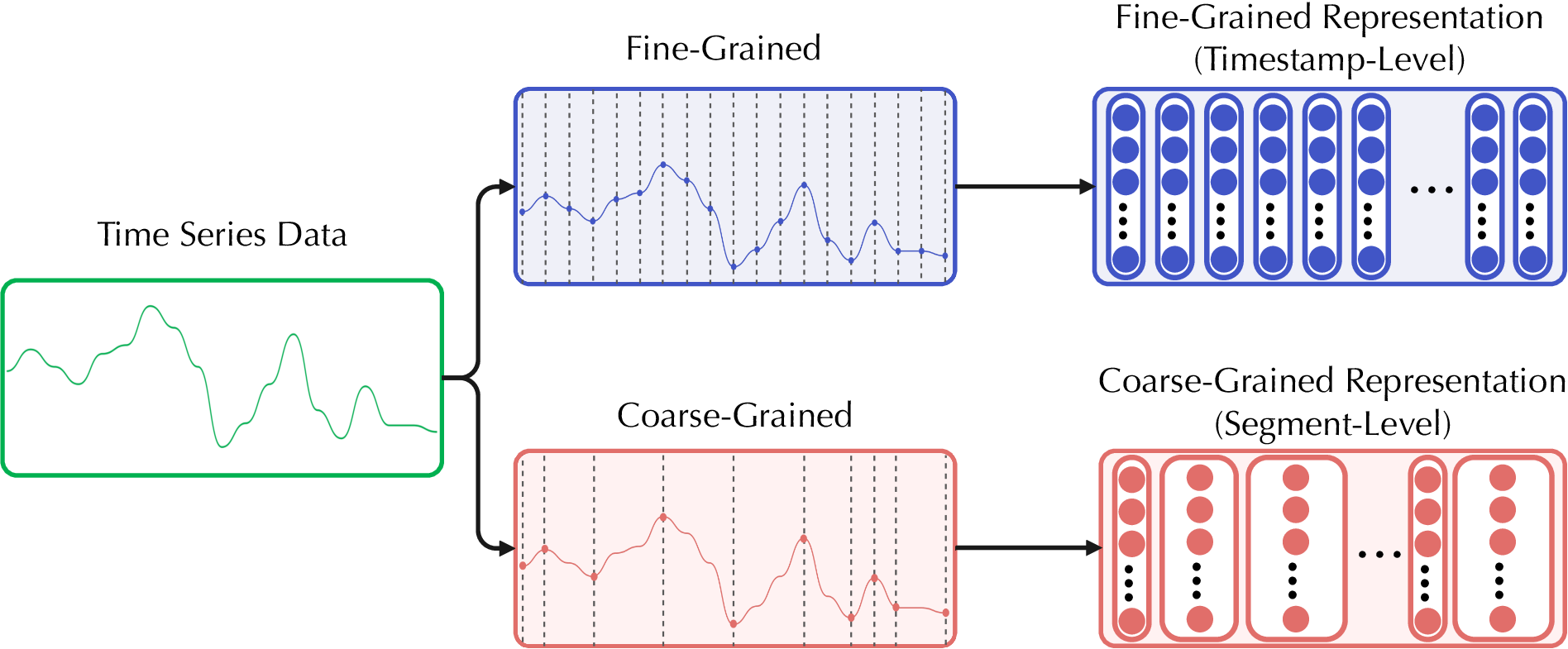}
\caption{Main differences between fine-grained representation learning and coarse-grained representation learning of time series.} \label{fig2}
\end{figure}

Fine-grained representation, i.e., timestamp-level representation learning, is the most traditional concept for the representation learning of time series. The objective of this method is to capture the relationships and dependencies between the different dimensions of the time series data at each point in time. Time2Vec (T2V) \cite{kazemi2019time2vec} is a typical timestamp-level representation learning method developed to capture temporal patterns and dependencies within the data. This method is based on Word2Vec \cite{church2017word2vec}. However, T2V may require detailed hyperparameter tuning to achieve an optimal performance. Selecting the appropriate dimensions for continuous vector representations, an appropriate learning rate, and determining the appropriate context window size can be challenging and time-consuming. Compared with T2V, The Time Series Transformer (TST) model \cite{zerveas2021transformer} provides more advantages. The TST model is a deep learning-based approach for time series analysis that leverages the transformer architecture \cite{vaswani2017attention}, originally designed for NLP tasks. The transformer architecture is known for its self-attention mechanism \cite{zhao2020exploring}, which can capture complex dependencies and patterns within sequences. The TST model can be used for various time series tasks, such as forecasting, classification, anomaly detection, and feature extraction.

Coarse-grained representation is referred to as as segment-level representation learning, i.e., learning representations for segments or subseries within an entire time series. These methods are focused on capturing global patterns and long-range dependencies in time series data, which can be beneficial for various tasks wherein the focus is on understanding local patterns and range dependencies in the data. The symbolic aggregate approximation (SAX)-based method \cite{yu2019novel} is a widely-used method for time series data representation and dimensionality reduction. In particular, it converts a continuous-valued time series into a discrete, symbolic representation while preserving the essential shape and trends of the original data. The SAX-based method can reduce the storage requirements with lower computational complexity. Additionally, the SAX-based method can be readily extended or combined with other techniques, such as indexable SAX (iSAX) \cite{shieh2009sax} or multivariate SAX (MSAX) \cite{anacleto2020msax}. However, the dimensional reduction and discretization process of the SAX-based method may result in information loss. The Shapelet-based methods, such as ShapeNet \cite{li2021shapenet}, may be the most advanced segment-level representation learning method. These techniques are focused on identification of discriminant sub-sequences in time series data, which can be useful for tasks such as classification and anomaly detection. However, the computational complexity of shapelet discovery can be high, particularly for large datasets and long time series. 

\subsection{Multi-Granularity Representation Methods for Time Series}

Both coarse- and fine-grained representation learning have advantages and applicability scenarios that render them suitable for different types of time series analysis tasks. Within this context, the majority of existing studies were focused on a single granularity, and methods are developed based on a specific level of detail in time series data with the objective of predicting the labels corresponding to the granularity. 

However, in general, selecting the appropriate granularity for different tasks is a challenge that significantly depends on experience. Multi-granularity representations allow for information to be obtained from various perspectives within time series data, thus providing a more comprehensive understanding of the underlying patterns and structures. For example, in the analysis of stock market data, fine-grained representation learning methods can analyze high-frequency data such as intraday price movements. This helps to identify short-term trends and patterns. Coarse-granularity representations, such as daily or weekly price movements, can be useful for identifying long-term trends and patterns in the stock market, such as the overall market direction, support and resistance levels, and seasonal trends. Therefore, an increasing number of previous studies \cite{hou2021stock} \cite{yang2023multi} were focused tend on multi-granularity representation learning.  

The multi-granularity substructure-aware representation learning algorithm for time series (MS-SRALAT \cite{boonchoo2022ms}) is an advanced semantic representation of a symbol sequence that is generated corresponding to a time series by an approximation algorithm that can capture the structure of the original data. In particular, it is a quite concise and easily implementable method that utilizes the SAX and produces the representation of a time series by transforming the target time series into an SAX sentence and aggregating those embeddings of the SAX words in the SAX sentences. However, the SAX information in this framework cannot reveal meaningful semantic information, which limits its performance.

\section{Methodology}
\subsection{Overview}

\begin{figure}[!b]
\includegraphics[width=\textwidth]{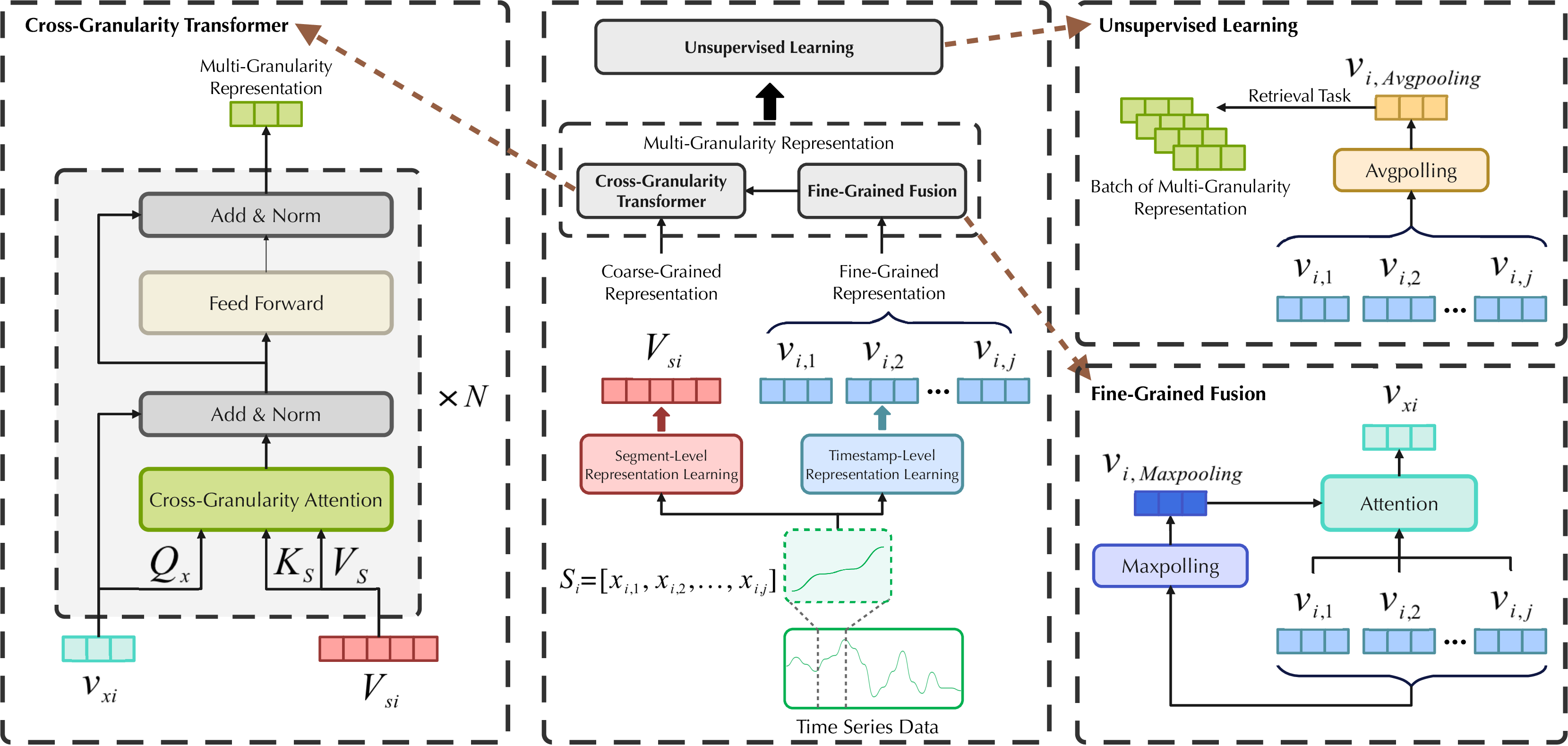}
\caption{\textbf{Middle:} The structure of unsupervised multi-granularity representation learning for time series. \textbf{Left:} Details of cross-granularity transformer. \textbf{Right:} Details of the fine-grained fusion and retrieval-based unsupervised learning.} \label{fig3}
\end{figure}

This section presents the proposed MUG framework and the relevant algorithms are described. The structure of the MUG is shown in Fig. 3. Each training sample $X\in \mathbb{R}^{w\times m}$, which is a time series of length $w$ and $m$ different variables, constitutes a sequence of $w$ time series $x_t \in \mathbb{R}^m : X \in \mathbb{R}^{w\times m} = [x_1, x_2,..., x_w]$. Moreover, for each segment $S_i \in X$ in the time series, $S_i = [x_{i,1}, x_{i,2},..., x_{i,j}]$, which implies that segment $S_i$ has $j$ timestamp points in the time series. 

First, for each segment $S_i$, the proposed framework employs two different representation learning algorithms for the coarse- and fine-grained time series data, thus constructing two distinct feature vectors. Using both granularities, the objectives of the model is to capture the different levels of the information present in the time series data, thus providing a more comprehensive representation. Fine-grained representations focus on local patterns and detailed information within the data, whereas coarse-grained representations capture the high-level patterns and global structures in the data. Thereafter, for the fusion of fine-grained representation of time series, a variant of the attention mechanisms was employed to combined the features of each timestamp-level representation of the time series, and generate a more comprehensive representation vector to represent the fine-grained information in certain segments of the time series. Moreover, for coarse-grained representations, a cross-granularity transformer with cross-granularity attention mechanism was employed to map coarse-grained representations onto fine-grained representations. Finally, with focus on the demand for unsupervised learning in multi-granularity representation learning, a retrieval-based task was selected as the training task for unsupervised learning. Based on the characteristics of the retrieval task, a novel loss function was designed to improve the performance of the training model. 

\subsection{Fine-Grained Fusion}

The structure of the fine-grained fusion is shown at the bottom right of Fig. 3. This part is based on a variant of the attention mechanism, which was first designed as an NLP model for multi-granularity relation extraction \cite{nie2018mention}. This type of attention mechanism helps the model combine the feature information from each inputs, which is suitable for the multi-granularity representation learning framework, in the stage of representing the comprehensive feature vector of the fine-grained representation learning of time series.  

Based on timestamp-level representation learning methods, the values of timestamp points can be embedded into a fine-grained representation, which can be formalized using Equation (1). 

\begin{equation}
v_i = \{v_{i,1}, v_{i,2},..., v_{i,j}\} =f_{encoder}(x_{i,1}, x_{i,2},..., x_{i,j})
\end{equation}
Where $v_i = {v_{i,1}, v_{i,2},..., v_{i,j}}$ are the representation vectors of timestamp-level inputs. Index $i$ indicates that these timestamp points are from Segment $S_i$ in the time series. 

Moreover, as in the original research, to built a comprehensive representation vector without any external information, a maximum pooling operation should be employed to obtain the shallow features of each timestamp-level inputs. $v_{i, Maxpooling} = Maxpooling(v_i)$. 

Thereafter, to capture the comprehensive information of the fine-grained representations inputs of the time series, the fine-grained fusion part combined the timestamp-level feature and the maximum pooling value of each timestamp-level inputs. Specifically, the maximum pooling representation of these timestamp points can be used as the Query vector in the attention mechanism to obtain the fusion feature of fine-grained representation by Equation (2).

\begin{equation}
v_{xi} = Sfotmax\left(\frac{Q\cdot K^T}{\sqrt{d}}\right) \cdot V= Softmax\left(\frac{v_{i, Maxpooling}\cdot v_i}{\sqrt{d}}\right) \cdot v_i
\end{equation}
Where $d$ denotes the dimension of the representation vector and is used to normalize the vectors. In the remainder of this paper, $d_{(\cdot)}$ is used to represent the dimension of representation vector. 

After the fine-grained fusion, the comprehensive representation vector of fine-grained representation learning is computed, which is employed to calculate multi-granularity representation in the subsequent steps.

\subsection{Cross-Granularity Transformer}

Cross-granularity representation is the subsequent step in the proposed framework. Unlike the fusion of fine-grained representation learning to obtain a comprehensive vector, cross-granularity representation has its own challenge. 

Cross-granularity representation, which refers to the combination of coarse- and fine-grained information in a unified framework, are generally subject to redundancy. There may be overlapping or redundant information between the different granularities, thus leading to inefficiencies in the representation and potential over-fitting in the learning process. Additionally, complexity is a critical issue. Combining features from different granularities increases the complexity of the model, thus potentially increasing the computational requirements and training time. In addition, determining the optimal method for the fusion or integration of features from different granularities to generate a cohesive representation that effectively captures the underlying patterns in the data can be challenging. Therefore, multi-granularity feature fusion has attracted significant attention with respect to multi-granularity representation.

\begin{figure}[!b]
\includegraphics[width=\textwidth]{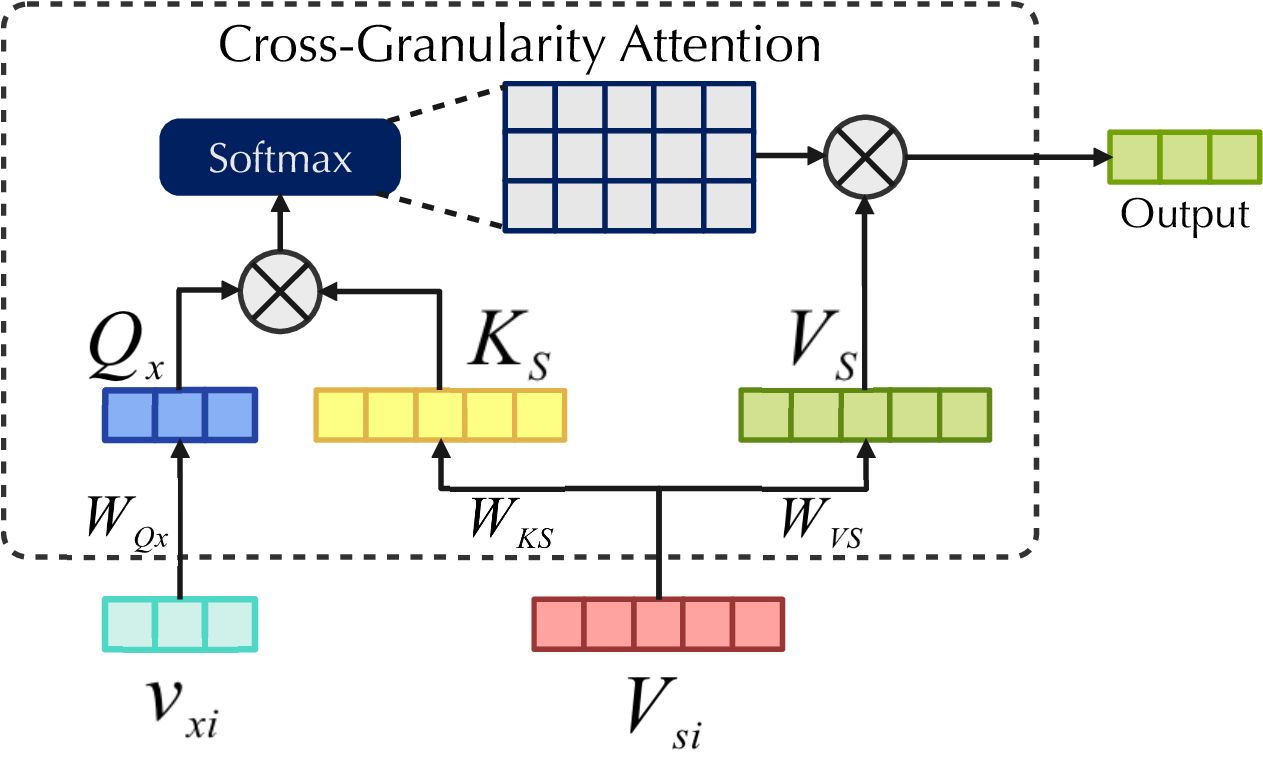}
\caption{Structure of cross-granularity attention mechanism.} \label{fig4}
\end{figure}

As mentioned previously, most existing models primarily focus on the simple fusion of decision results and generally require the re-design of representation models. Consequently, they cannot utilize existing, well-performing representation methods and lack the flexibility to adapt to different scenarios. To address this issue and more extensively utilize various existing excellent time series representation learning methods, we designed a cross-granularity transformer architecture based on the cross-granularity attention mechanism. The structure of the cross-granularity attention mechanism is shown in Fig. 4.

To introduce the cross-granularity attention mechanism, we considered two representation vectors, namely,  $v_{xi}$ and $V_{Si}$ from fine- and coarse-grained representation learning, respectively, where $v_{xi} \in \mathbb{R}^{d_x}$ and $V_{Si}\in \mathbb{R}^{d_S}$. Based on the transformer architecture of multi-modal data fusion \cite{tsai2019multimodal}, we hypothesized that a suitable method for the fusion of cross-granularity information is to provide a latent adaptation across multi-granularity. In the proposed framework, it means $V_{Si}$ to $v_{xi}$ (coarse- to fine-grained). 

We defined the Query as $Q_x = v_{xi}W_{Q_x}$, which is a linear transformation of the fine-grained representation input, the Key as $K_S = v_{Si}W_{K_S}$ and the Value as $V_S = v_{Si}W_{V_S}$, which are the line transformations of the coarse-grained representation input. Moreover, $W_{Q_x}$, $W_{K_S}$ and $W_{V_S}$ are the weights. The latent adaptation from $V_{Si}$ to $v_{xi}$ is presented as the cross-granularity attention as follows:

\begin{equation}
Attention(Q_x,K_S,V_S)=softmax\left(\frac{Q_xK_S^T}{\sqrt{d_k}}\right)V_S
\end{equation}

The output of the cross-granularity attention mechanism has the same length as $v_{xi}$. Using this equation, the mapping of coarse- onto fine-grained representations was established.

\subsection{Unsupervised Learning}

The primary objective of unsupervised learning is to learn useful features or representations from the data without using any labeled information. This can be particularly beneficial for time series data analysis because obtaining labeled data can be time-consuming and expensive. 

However, the design of algorithm for unsupervised learning is challenging. Several of these difficulties can be attributed to a lack of labeled data, which indicates that the model should identify the underlying structure and relationships within the data without any explicit guidance. However, constructing positive and negative sample pairs for unsupervised training is difficult. The construction of sample pairs is closely related to the selection of the unsupervised training tasks. In this framework, we designed an unsupervised training tasks based on retrieval task. 

There were several reasons for us for selecting retrieval task. First, unlike other unsupervised learning models, the proposed MUG is required to accomplish multi-granularity representation vector fusion during the training process, which indicates that constructing positive and negative sample pairs before training is different. The representation vectors of the positive and negative samples are not in the same vector space as the anchor. Therefore, traditional unsupervised contrastive learning methods based on similarity measures cannot be applied in this scenario. Constructing positive and negative sample pairs during the training process is undoubtedly complex and time consuming. In addition, traditional loss functions are subject to several limitations. If the selected triplets are not informative, the triplet loss may rapidly converge to zero, thus leading to the degradation of the model performance. 

To solve these issues, we designed an unsupervised learning method using a retrieval training task and applied a novel loss function in the training (as shown at the top right of Fig. 3). First, because we used the maximum pooling method in fine-grained fusion, we applied the average Pooling (Avgpooling) method to construct the query vector in the retrieval task. Thereafter, the multi-granularity representation corresponds to the query vector, in addition to other randomly selected multi-granularity representation vectors form the query object together. Assuming that the query vector is $y_q$, the correct multi-granularity representation corresponding to the query vector is $y_t$, and the other multi-granularity representation vectors are $y_j$. The ranking of $y_t$ can be expressed as follows:

\begin{equation}
1+\sum I \left( \left\|H(y_q)-H(y_t)\right\| \ge \left\|H(y_q)-H(y_j)\right\| \right)
\end{equation}
where $H(Z)$ is a representation vector value of $Z$, and $I (a \ge b) $ is a function that is transformed to $1$ when $a \ge b$ and $0$ in other cases. In the above expression, ranking is used to describe the relative distances. 

To convert the ranking situation into a similarity metric that can be used as a loss function, the Spearman correlation coefficient \cite{de2016comparing} was used to calculate the similarity. The Spearman correlation coefficient is a statistical measure that evaluates the strength and direction of the monotonic relationship between two variables. The equation of the Spearman correlation coefficient of $y_t$ can be expressed as follows:

\begin{equation}
Similarity_t = \frac{n-\pi_t}{n-1}
\end{equation}
where $\pi_t$ denotes the ranking of $y_t$. Using the Spearman correlation coefficient, the relative similarity was used in the loss function to accelerate model convergence and improve model accuracy. Moreover, the traditional loss function is applicable in such cases.

We did not use ranking losses \cite{kemertas2020rankmi} because we found that the binary classification loss \cite{sypherd2019tunable} demonstrated a superior performance, which was similar to that reported in \cite{qi2020imagebert}. The equation of binary classification loss function is expressed as follows:

\begin{equation}
\mathcal{L}_{BCE} = -[ylog(\theta)+(1-y)log(1-\theta)]
\end{equation}

where the ground truth labels $y \in (0,1)$ and $\theta$ represent the similarity. 

Therefore, by combining Equation 5 and 6, the novel loss function can be expressed as follows: 

\begin{equation}
\mathcal{L}_{BCE} = -[ylog(\frac{n-\pi_t}{n-1})+(1-y)log(1-\frac{n-\pi_t}{n-1})]
\end{equation}

\section{Results and discussion}

As detailed in this section, we tested the effectiveness of the proposed framework by analyzing its performance on classification task, which was used as a downstream task, to prove the effectiveness of the proposed multi-granularity representation learning framework. Moreover, to highlight the advantages in real-world time series data, comparative experiments was conducted with other multi-granularity representation methods under simulated real-world scenario. Additionally, a case study was conducted to recall the example introduced in Section 1. 

\subsection{Classification}
In the classification task, the output multi-granularity representation vector of the proposed framework was passed through a SoftMax function to obtain a distribution over the classes. TST is used with ShapeNet, which are introduced in Section 2.1, as the fine- and coarse-grained representation parts in our framework. In this task, we demonstrated that the proposed framework demonstrated a superior performance to those of other non-deep learning method and unsupervised methods. 

We used the following ten multivariate datasets from the UEA time series classification archives \cite{bagnall2018uea}, which provided multiple datasets from different domains with varying dimensions, unequal lengths, and missing values. We selected datasets from a diverse range of domains across science and engineering from Monash University, UEA \& UCR Time Series Classification Repository. Selection was made to ensure diversity with respect to the dimensionality and length of the time series samples, in addition to the number of samples and classes (when applicable). Furthermore, we included both the "easy" and "difficult" datasets, where the baselines performance were significantly high or low, respectively. A summary of these datasets is provided in Table 1. 

\begin{table}[!tb]\centering
\caption{Summary of UEA multivariate datasets.}\label{tab1}
\setlength{\tabcolsep}{2mm}{
\begin{tabular}{c|ccccc}
\hline\hline
Dataset & Train Size & Test Size & Length & Classes & Dimensions \\
\hline
EthanolConcentration &  261 & 263 & 1751 & 4 & 3 \\
FaceDetection & 5890 & 3524 & 62 & 2 & 144 \\
Handwriting &  150 & 850 & 152 & 26 & 3 \\
Heartbeat &  204 & 205 & 405 & 2 & 61 \\
JapaneseVowels &  270 & 370 & 29 & 9 & 12 \\
PEMS-SF &  267 & 173 & 144 & 7 & 983 \\
SelfRegulationSCP1 & 268 & 293 & 896 & 2 & 6 \\
SelfRegulationSCP2 & 200 & 180 & 1152 & 2 & 7 \\
SpokenArabicDigits &  6599 & 2199 & 93 & 10 & 13 \\
UWaveGestureLibrary & 2238 & 2241 & 315 & 8 & 3 \\
\hline\hline
\end{tabular}}
\end{table}

The UEA archives provides an initial benchmark for existing models with accurate baseline information. Based on the performance metrics provided in the UEA archives, we selected the following three models as our baselines:

\begin{itemize}
\item[-]Dimension-dependent dynamic time warping (DTW\_D) \cite{chen2013dtw}: it uses a weighted combination of raw series and first-order differences for the neural network classification with either Euclidean distance or full-window dynamic time warping (DTW). Additionally, it develops the traditional DTW method and suits every data series.
\item[-]ROCKET \cite{dempster2020rocket}: it is based on a random convolutional kernel similar to a shallow convolutional neural network. It can achieve rapid and accurate time series classification using random convolutional kernels. 
\item[-] Long short-term memory (LSTM) model \cite{karim2017lstm}: it is a type of recurrent neural network (RNN) architecture that is designed to overcome the limitations of traditional RNNs in capturing long-term dependencies in sequential data. 
\end{itemize}

\begin{table}[!tb]\centering
\caption{Accuracy results of the proposed and other methods.}\label{tab2}
\setlength{\tabcolsep}{0.5mm}{
\begin{tabular}{c|cccccc}
\hline\hline
Dataset & \makecell[c]{MUG\\(TST-ShapeNet)} & TST & ShapeNet & DTW\_D & ROCKET & LSTM\\
\hline
EthanolConcentration & {\bfseries 0.471} & 0.337 & 0.312 & 0.323 & 0.452 & 0.323\\
FaceDetection & {\bfseries 0.694}& 0.681 & 0.602 & 0.529 & 0.647 & 0.577 \\
Handwriting &  0.366 & 0.305 & 0.451 & 0.286 & {\bfseries 0.588} & 0.152\\
Heartbeat &  {\bfseries 0.780} & 0.776 & 0.756 & 0.717 & 0.756 & 0.722\\
JapaneseVowels & {\bfseries 0.997} &  0.994 & 0.984 & 0.949 & 0.962 & 0.797\\
PEMS-SF & {\bfseries 0.919} & {\bfseries 0.919} & 0.751 & 0.711 & 0.751 & 0.399\\
SelfRegulationSCP1 & {\bfseries 0.945} & 0.925 & 0.782 & 0.775 & 0.908 & 0.689 \\
SelfRegulationSCP2 & {\bfseries 0.615} & 0.589 & 0.578 & 0.539 & 0.533 & 0.466\\
SpokenArabicDigits & {\bfseries0.995} & 0.993 & 0.975 & 0.963 & 0.712 & 0.319\\
UWaveGestureLibrary & 0.905 & 0.903 & 0.906 & 0.903 & {\bfseries0.944} & 0.412 \\
\hline
Average Accuracy & {\bfseries0.768} & 0.742 & 0.710 & 0.669 & 0.723 & 0.486 \\
Average Rank & {\bfseries1.4} & 2.4 & 3.1 & 4.5 & 2.9 & 5.3 \\
\hline\hline
\end{tabular}}
\end{table}

Table 2 presents the classification results for the time series, where bold values indicate the optimal values. As shown in Table 2, the proposed framework demonstrated the highest performance on eight of the ten datasets, thus achieving an average rank of 1.4$^{th}$, followed by TST, which demonstrated one highest performance with average ranked 2.4$^{th}$. ROCKET, which demonstrated the optimal performances for the remaining two datasets, and on average, was ranked 2.9$^{th}$. From the data presented in the table, it can be concluded that the effectiveness of proposed framework significantly increased as the amount of data increased. In addition, comparing the performances of MUG (TST-ShapeNet), TST and ShapeNet, it is evident that multi-granularity representation can achieve a superior performance to that of the single-granularity representation method. This is because multi-granularity methods can capture complex temporal dependencies and patterns that may be presented at different scales or resolutions in the data.

\subsection{Comparative Experiments}

This section presents a discussion on the performance of proposed MUG and other multi-granularity models with respect to real-world time series data. Accordingly, we used several UCR archives \cite{dau2019ucr} and randomly added Gaussian noise and segments of time series data from other classes to the time series data to simulate the real-world cases analyzed in Section 1. A summary of these datasets is provided in Table 3. 

\begin{table}[!tb]\centering
\caption{Summary of simulated real-world time series data from the UCR datasets.}\label{tab3}
\setlength{\tabcolsep}{2mm}{
\begin{tabular}{c|ccccc}
\hline\hline
Dataset & Train Size & Test Size & Length & Classes \\
\hline
Adiac &  390 & 391 & 200 & 37  \\
Beef & 30 & 30 & 500 & 5  \\
Fish &  175 & 175 & 480 & 7  \\
Gun-Point &  50 & 150 & 170 & 2  \\
CBF &  30 & 900 & 160 & 3  \\
Trace &  100 & 100 & 300 & 4  \\
\hline\hline
\end{tabular}}
\end{table}

It should be noted that the lengths of these six datasets are longer than the original lengths in UCR archives, which is due to the simulation of the real-world situations. 

For comparison, we selected the MS-SRALAT framework introduced in Section 2.2. The ROCKET algorithm was used as a control group to further illustrate the performance difference between the single-granularity and multi-granularity methods. The results of the comparative experiments are presented in Table 4.

\begin{table}[!b]\centering
\caption{Accuracy comparison between single-granularity and multi-granularity methods.}\label{tab4}
\begin{tabular}{c|ccc}
\hline\hline
Dataset & MUG(TST-ShapeNet) & MS-SRALAT & ROCKET \\
\hline
Adiac &  0.435 & 0.379 & {\bfseries0.468}   \\
Beef & {\bfseries0.614} & 0.550 & 0.458  \\
Fish & {\bfseries0.758} & 0.671 & 0.469  \\
Gun-Point &  {\bfseries0.859} & 0.701 & 0.647  \\
CBF &  0.900 & {\bfseries0.934} & 0.887  \\
Trace &  {\bfseries0.877} & 0.860 & 0.713  \\
\hline\hline
\end{tabular}
\end{table}

By analyzing the results in Table 4, the proposed framework obtained superior results to those obtained by MS-SRALAT. This is because of the more advanced fine- and coarse-grained representations selected in the proposed framework, in addition to the more fixable structure and improved fusion method. The proposed framework can therefore improve the accuracy of downstream tasks.

\subsection{Case Study}

\begin{table}[!t]\centering
\caption{Summary of ECG200 and TwoLeadECG.}\label{tab5}
\setlength{\tabcolsep}{2.6mm}{
\begin{tabular}{c|ccccc}
\hline\hline
Dataset & Train Size & Test Size & Length & Classes \\
\hline
ECG200 &  100 & 100 & 96 & 2 \\
TwoLeadECG &  23 & 1139 & 82 & 2 \\
Combined ECG &  123 & 1239 & 82 & 2 \\
\hline\hline
\end{tabular}}
\end{table}

In this case study, the example in Section 1 was considered more comprehensively to clarify the motivation for the study. As introduced in Section 1, by analyzing the characteristic of real-world ECG data, it can be concluded that the complex temporal dependencies and patterns may exist at different scales or resolutions in the data. This section presents experimental design to further address this issue. 

To simulate real-world ECG data, we combined two other datasets from the same subject obtained from various sources. Specifically, we used the ECG 200 and TwoLeadECG as datasets from the UCR time series classification archive. Both datasets traced the recorded electrical activity and contained two classes: normal heartbeats and myocardial infarctions (MIs). We randomly combined these two datasets and reshaped the length of the combined ECG dataset to obtain a regular time series. The details of these datasets are presented in Table 5. 

We selected DTW\_D and Shapelet Transform \cite{arul2021applications}  as the baseline algorithms. The Shapelet Transform (ST) is based on the shapelet method, which represent fine- and coarse-grained representations respectively. First, separate experiments were conducted using these two datasets. We then conducted an experiment using the combined dataset. The experimental results of this case study are listed in Table 6.

\begin{table}[!tb]\centering
\caption{Accuracy results of proposed and other methods.}\label{tab6}
\setlength{\tabcolsep}{5mm}{
\begin{tabular}{c|ccc}
\hline\hline
Dataset & MUG(TST-ShapeNet) & DTW\_D  & ST\\
\hline
ECG200 & {\bfseries 0.930}  & 0.880 & 0.840\\
TwoLeadECG &  {\bfseries 0.993} & 0.868 & 0.984\\
Combined ECG &  {\bfseries 0.800} & 0.442 & 0.510\\
\hline\hline
\end{tabular}}
\end{table}

\section{Conclusions}

In this study, we investigated the significance of exploring multi-granularity patterns for time series representation learning and proposed a multi-granularity framework for the unsupervised representation learning of time series. In particular, this paper proposes a novel unsupervised learning framework to build association between timestamp-level and segment-level features. To address the loss function issue in multi-granularity representation learning, a retrieval task for time series data with a special loss function was also designed. Experiments on public datasets and real-world data demonstrated the effectiveness of our MUG framework. In the future, we plan to employ more fine- and coarse-grained representation models in the proposed MUG framework to discuss the generality across different multiple granularity models. Moreover, we will focus on developing a more general framework that can combined more than two multi-granularity representation methods.

%
%
%
%
%
%

\end{document}